\documentclass{article}

\usepackage{PRIMEarxiv}

\usepackage{authblk}
\usepackage[utf8]{inputenc} 
\usepackage[T1]{fontenc}    
\usepackage{hyperref}       
\usepackage{url}            
\usepackage{booktabs}       
\usepackage{amsfonts}       
\usepackage{nicefrac}       
\usepackage{microtype}      
\usepackage{lipsum}
\usepackage{fancyhdr}       
\usepackage{graphicx}       
\usepackage{amsmath}
\usepackage{amssymb}
\usepackage{comment}
\usepackage{multirow}
\usepackage[numbers]{natbib}
\usepackage{float}
\usepackage{subfig} 
\usepackage[table]{xcolor}
\definecolor{bestblue}{RGB}{180, 210, 245} 
\definecolor{secondblue}{RGB}{224, 238, 255} 
\graphicspath{{media/}}     

\title{Shallow Preference Signals: Large Language Model Aligns Even Better with Truncated Data?
}

\begin{document}

\author[2]{Xuan Qi$^*$}
\author[1]{Jiahao Qiu$^*$}
\author[3]{Xinzhe Juan}
\author[1]{Yue Wu$^\dagger$}
\author[1]{Mengdi Wang$^\dagger$}

\affil[1]{AI Lab, Princeton University}
\affil[2]{IIIS, Tsinghua University}
\affil[3]{Department of Computer Science \& Engineering, University of Michigan}

\maketitle
\begingroup
\let\thefootnote\relax
\footnotetext{$^*$ Equal contribution.}
\footnotetext{$^\dagger$ Correspondence to: \texttt{frankwupku@gmail.com}, \texttt{mengdiw@princeton.edu}.}
\endgroup

\begin{abstract}
Aligning large language models (LLMs) with human preferences remains a key challenge in AI. Preference-based optimization methods, such as Reinforcement Learning with Human Feedback (RLHF) and Direct Preference Optimization (DPO), rely on human-annotated datasets to improve alignment. In this work, we identify a crucial property of the existing learning method: the distinguishing signal obtained in preferred responses is often concentrated in the early tokens. We refer to this as \textit{shallow preference signals}.

To explore this property, we systematically truncate preference datasets at various points and train both reward models and DPO models on the truncated data. \textbf{Surprisingly}, models trained on truncated datasets, retaining only the first half or fewer tokens, achieve comparable or even superior performance to those trained on full datasets. For example, a reward model trained on the Skywork-Reward-Preference-80K-v0.2 dataset outperforms the full dataset when trained on a 40\% truncated dataset. This pattern is consistent across multiple datasets, suggesting the widespread presence of \textit{shallow preference signals}.

We further investigate the distribution of the reward signal through decoding strategies. We consider two simple decoding strategies motivated by the shallow reward signal observation, namely {Length Control Decoding} and {KL Threshold Control Decoding}, which leverage shallow preference signals to optimize the trade-off between alignment and computational efficiency. The performance is even better, which again validates our hypothesis.

The phenomenon of \textit{shallow preference signals} highlights potential issues in LLM alignment: existing alignment methods often focus on aligning \textbf{only the initial tokens} of responses, rather than considering the full response. This could lead to discrepancies with real-world human preferences, resulting in suboptimal alignment performance. Our code is available at \url{https://github.com/THUQiXuan/Shallow-preference-signal}.
\end{abstract}

\section{Introduction} 
\label{sec: introduction}

Aligning large language models (LLMs) with human preferences is a core challenge in artificial intelligence (AI) research \cite{wang2023aligning}. Preference datasets \cite{DBLP:journals/corr/abs-2410-18451, DBLP:journals/corr/abs-2310-01377, DBLP:journals/corr/abs-2112-00861, DBLP:journals/corr/abs-2204-05862} have played a critical role in addressing this challenge by capturing human judgments of model outputs. These datasets enable the identification and prioritization of responses that are more aligned with human expectations. Preference-based optimization techniques, such as Reinforcement Learning with Human Feedback (RLHF) \cite{DBLP:conf/nips/Ouyang0JAWMZASR22} and Direct Preference Optimization (DPO) \cite{DBLP:conf/nips/RafailovSMMEF23}, rely on these datasets to refine the decision-making process of models.

Despite the promise of these methods, there are several challenges associated with them. Recent work \cite{DBLP:journals/corr/abs-2409-11704, DBLP:conf/emnlp/ParkJRKC24, DBLP:conf/acl/ParkREF24, DBLP:journals/corr/abs-2402-01306} has highlighted that reward models trained using RLHF may suffer from reward hacking. Factors such as response format, length, and even the inclusion of emojis can influence quality judgments, resulting in potential inaccuracies. In this paper, we introduce a previously underexplored aspect of preference data. Specifically, we observe that the signal indicating the superiority of the chosen response over the rejected one is not uniformly distributed across the entire response. In many cases, the relative quality of responses can be determined from only the early portion of the response—or even just a few tokens—rather than requiring an evaluation of the entire response. We refer to this phenomenon as \textbf{shallow preference signals}. This observation suggests that preference-based optimization methods may not need to rely on the full response to effectively capture the distinguishing features of higher-quality responses. 

\begin{figure*}[ht!]
    \centering
    \includegraphics[width=1.0\textwidth]{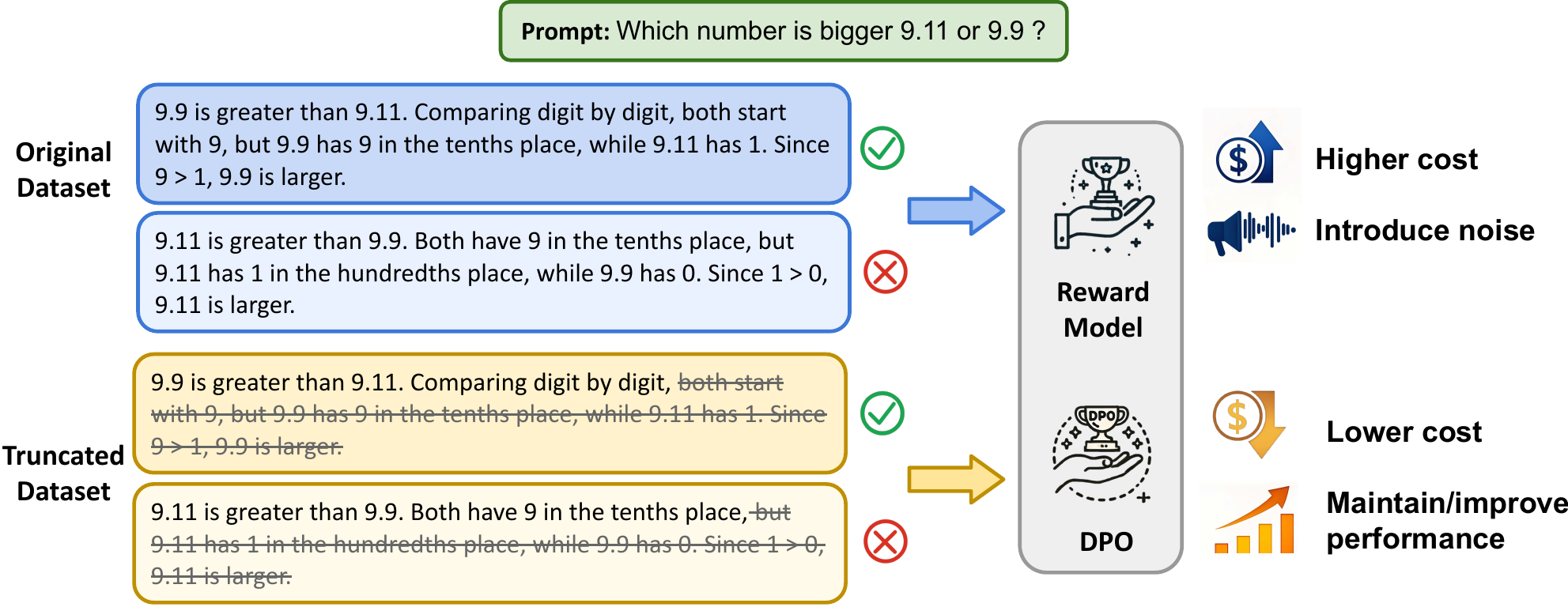}
    \caption{An example illustrating the phenomenon of shallow preference signals. It demonstrates how the relative quality of two responses can be determined from the early portion of the response, or even from the first sentence. Training with only the initial part allows the model to capture most of the preference signals while conserving resources.}
    \label{fig:figure1}
\end{figure*}

We hypothesize that focusing on the early portion of the response allows models to capture the most salient preference signals, resulting in more efficient training and potentially improved alignment performance. To test this hypothesis, we introduce a methodology where preference data is truncated at various positions, and models are trained on these truncated datasets. We analyze the distribution of preference signals in response pairs and conduct systematic experiments to validate the hypothesis that models trained on truncated preference data perform comparably to models trained on the full dataset. This is confirmed for both reward models and models fine-tuned with DPO. Our findings demonstrate that the distinguishing features between the chosen and rejected responses are concentrated in the early part of the response. In fact, models trained on truncated datasets—using only the first half or fewer tokens of each response—achieve similar, or even superior, performance compared to those trained on the full dataset. For instance, a reward model trained on the Skywork-Reward-Preference-80K-v0.2~\cite{DBLP:journals/corr/abs-2410-18451} dataset achieves an accuracy of only 75.85\% on RewardBench~\cite{DBLP:journals/corr/abs-2403-13787}. However, when the dataset is truncated to 50\% and 40\%, the accuracy increases to 75.88\% and 76.35\%, respectively. Even with a truncation to 25\%, the accuracy remains at 69.92\%. Similarly, a reward model trained on the RLHFlow-pair-data-v2-80K-wsafetyRLHFlow-pair-data-v2-80K-wsafety\footnote{\url{https://huggingface.co/datasets/RLHFlow/pair_data_v2_80K_wsafety}} dataset achieves an accuracy of 65.96\% on RewardBench. After truncating the dataset to 50\% and 40\%, the accuracy improves to 72.16\% and 69.71\%, respectively, with accuracy remaining at 62.44\% for a 33\% truncation.

Furthermore, our experiments suggest that the shallow preference signal phenomenon significantly impacts LLM content generation. Based on this observation, we find that simple strategies can perform well without needing complex decoding approaches. Recent work \cite{DBLP:journals/tmlr/YangLCP024, DBLP:journals/corr/abs-2402-16844, DBLP:journals/corr/abs-2411-10666, DBLP:journals/corr/abs-2407-01955} has proposed various decoding strategies, but our findings indicate that by focusing on the early portion of the response, we can achieve an optimal trade-off between reward and KL divergence. To test this, we explore two decoding strategies—{Length Control Decoding} and {KL Threshold Control Decoding}—to see if the early-token bias observed during training affects generation at inference time. Our results show that the differences between the DPO model trained on full preference data and the reference model are most noticeable in the early tokens of the generated response. As more of the response is generated, the difference decreases. This suggests that the reward signal in DPO training is concentrated in the early tokens, rather than being evenly distributed. \cite{DBLP:conf/iclr/LinRLDSCB024} also explores token distribution differences between base LLMs and aligned models, though their method primarily focuses on in-context learning, avoiding parameter fine-tuning.

Meanwhile, the findings of this paper may shed light on existing problems in LLM alignment. Our experiments validates that current alignment methods often focus on aligning earlier tokens, rather than considering full sentences. The latter portions of answers generated by LLM tend to be generated through an auto-regressive mechanism, which does not exhibit significant quality variation through our decoding experiments. Through extensive experiments, we validate our hypothesis that focusing on the early portion of the response allows models to capture the most salient preference signals, resulting in more efficient training and potentially improved alignment performance. \textbf{However, alignment with truncated data is shallow alignment which only improves the performance on metrics but may keep further away from the real-world alignment with human values.} \cite{DBLP:journals/corr/abs-2406-05946} proposes a related issue, but their work is confined to safety alignment and does not extend to the broader alignment challenges present in LLMs. Instead, our work validates the phenomenon more systemically and extensively.

In summary, the main contributions of our paper are as follows:
\begin{enumerate}
    \item We introduce and systematically validate the phenomenon of \textbf{shallow preference signals}, demonstrating that the distinguishing features between high-quality and low-quality responses are often concentrated in the early portion of the response.
    \item We show that training reward models and DPO models on truncated responses—using only the early portion—achieves performance comparable to or better than training on full responses. This finding holds across multiple datasets and supervision settings.
    \item We provide a new perspective on the limitations of current alignment pipelines. Specifically, we suggest that current alignment methods face the limitation of shallow alignment, emphasizing that alignment should go beyond just aligning a few tokens and consider full sentences for more effective results.
\end{enumerate}

\section{Related Works}
\label{sec: related_work}

\subsection{LLM Alignment with Human Preference}
Aligning the outputs of large language models with human preferences is a crucial problem in the field of LLMs~\cite{wang2023aligning}. One of the most notable advancements in this area is Reinforcement Learning from Human Feedback (RLHF)~\cite{DBLP:conf/nips/ChristianoLBMLA17, DBLP:conf/nips/Ouyang0JAWMZASR22}, which has led to the development of cutting-edge language models such as GPT-4o~\cite{DBLP:journals/corr/abs-2410-21276}, Gemini-2.0~\cite{DBLP:journals/corr/abs-2312-11805}, and Llama-3.1-70B-Instruct~\cite{DBLP:journals/corr/abs-2407-21783}. The traditional RLHF approach involves training a reward model to score the outputs of the language model, followed by fine-tuning using deep reinforcement learning algorithms like Proximal Policy Optimization (PPO)~\cite{DBLP:journals/corr/abs-2204-05862}. However, PPO faces challenges in alignment tasks due to its complexity, instability, and inefficiency~\cite{DBLP:conf/iclr/ChoshenFAA20,DBLP:journals/corr/abs-2005-12729}. Several works have sought to improve the RLHF paradigm from various angles in order to better align LLMs with human preferences~\cite{DBLP:journals/corr/abs-2305-10425,DBLP:conf/aistats/AzarGPMRVC24,DBLP:conf/icml/TangGZCMRRVPP24,DBLP:journals/corr/abs-2410-16033, chakraborty2024maxmin}. Among these, Direct Preference Optimization (DPO)~\cite{DBLP:conf/nips/RafailovSMMEF23} has gained significant attention, as it directly optimizes a policy using chosen and rejected pairs.

\subsection{Reward Model}
The reward model plays a critical role in RLHF~\cite{DBLP:conf/nips/ChristianoLBMLA17, DBLP:conf/nips/Ouyang0JAWMZASR22}. Traditional reward models are often assumed to follow a Bradley-Terry model~\cite{19ff28b9-64f9-3656-ba40-08326a05748e}, which provides a score for an entire output to indicate its preference~\cite{DBLP:journals/corr/abs-2310-00898,DBLP:conf/nips/ChristianoLBMLA17, DBLP:conf/nips/Ouyang0JAWMZASR22}. However, the Bradley-Terry model has limitations, particularly its inability to handle complex or intransitive preferences~\cite{DBLP:conf/icml/MunosVCARGTGMFM24,DBLP:conf/icml/SwamyDK0A24,DBLP:journals/corr/abs-2402-07314}. Some works have addressed this issue by discarding the Bradley-Terry assumption and instead modeling the probability that one response is preferred over another~\cite{DBLP:conf/acl/Jiang0L23,DBLP:conf/iclr/0002ZJKSLL24,DBLP:journals/corr/abs-2405-07863}. Additionally, other approaches have explored the construction of multi-objective reward models to capture human preferences more comprehensively~\cite{DBLP:journals/corr/abs-2307-09288,DBLP:conf/naacl/WangDZASEDSKSK24,DBLP:conf/emnlp/00030X0024}. Furthermore, some studies have proposed process reward models~\cite{DBLP:journals/corr/abs-2308-09583,DBLP:conf/iclr/LightmanKBEBLLS24,DBLP:journals/corr/abs-2410-11287} or step-wise reward models~\cite{DBLP:conf/icml/HavrillaRNDZHR24}, which have shown promising results, especially in reasoning tasks.

\subsection{Reward Hacking}
Reward hacking refers to the situation in which an agent or model optimizes a proxy reward that deviates from the true objective, leading to suboptimal or even undesirable behavior~\cite{DBLP:journals/corr/abs-2209-13085}. This phenomenon has been widely studied across various environments such as grid-worlds, Atari games, and text generation tasks~\cite{DBLP:conf/nips/Arjona-MedinaGW19, DBLP:conf/iclr/PanBS22, DBLP:conf/nips/XuPPBR0RG22}. Prior research has focused on categorizing different forms of reward hacking and developing mitigation strategies, such as regularizing policy optimization~\cite{laidlaw2024correlatedproxiesnewdefinition}, imposing a KL divergence penalty~\cite{DBLP:conf/nips/MiaoZ0BZT24}, and applying model merging techniques to either the policy or reward model~\cite{DBLP:journals/corr/abs-2409-11704}. Despite these efforts, existing approaches have notable limitations. In response, recent studies have introduced new definitions and strategies for mitigating reward hacking, including the concept of "hackability"~\cite{DBLP:journals/corr/abs-2209-13085} and the use of information-theoretic reward modeling~\cite{DBLP:conf/nips/MiaoZ0BZT24}. Furthermore, the application of reward hacking techniques to language models has been explored, particularly in improving the sample efficiency of preference learning~\cite{DBLP:journals/corr/abs-2412-16475}. In contrast to these prior approaches, our work mitigates a subset of reward hacking by truncating the model's responses and better aligning them with human preferences. This truncation process effectively reduces noise in the dataset, leading to improved accuracy. By removing certain noise components, our method can be seen as a novel approach to addressing reward hacking within the context of language models.

\section{Methodology}
\label{sec: methodology}

In this section, we introduce the methodology used to investigate and optimize the reward signal in large language models (LLMs) based on preference data. We first formalize the location of the reward signal within a response, followed by the approach of training reward models and Direct Preference Optimization (DPO) models using truncated preference datasets. Lastly, we describe the mixing strategy and the two novel decoding policies designed to improve model performance.

\subsection{Formulation of Reward Signal Location}

Consider a preference dataset containing pairs of responses, where one response is the \textit{chosen response} and the other is the \textit{rejected response}. The reward signal is defined as the inherent quality difference between these two responses. Let \( r_{\text{cho}}(i) \) denote the \textit{chosen response} for a given instance \( i \), and \( r_{\text{rej}}(i) \) denote the \textit{rejected response}. The objective is to model the reward signal \( R(i) \), which indicates the degree of preference for \( r_{\text{cho}}(i) \) over \( r_{\text{rej}}(i) \).

We hypothesize that the reward signal is concentrated in the early part of the response. To formalize this, let \( r_{\text{cho}}(i) = [y_1, y_2, \dots, y_T] \) and \( r_{\text{rej}}(i) = [z_1, z_2, \dots, z_T] \) represent the token sequences for the chosen and rejected responses, respectively, where \( T \) is the total number of tokens in each response. We define the reward signal at each token position \( t \) as the difference in the model’s log-probability for the chosen and rejected responses at that position:
$$
\begin{aligned}
R_t(i) &= \log p(y_t \mid x, y_{1:t-1}) - \log p(z_t \mid x, z_{1:t-1}),
\end{aligned}
$$
where \( x \) represents the input context, and \( \log p(y_t \mid x, y_{1:t-1}) \) is the log-probability of the token \( y_t \) in the chosen response at position \( t \), conditioned on the context \( x \) and the preceding tokens \( y_{1:t-1} \). Similarly, \( \log p(z_t \mid x, z_{1:t-1}) \) is the log-probability of the token \( z_t \) in the rejected response at the same position.

We argue that the total reward signal \( R(i) \) can be approximated as the cumulative sum of the reward signals up to a truncation point \( t_k \):
$$
\begin{aligned}
R(i) &= \sum_{t=1}^{t_k} R_t(i) =\log p(y_{1:t_k} \mid x) - \log p(z_{1:t_k} \mid x),
\end{aligned}
$$
where \( t_k \) represents the truncation point, beyond which the reward signal becomes less informative or introduces noise. This leads to the hypothesis that truncated responses up to position \( t_k \) preserve most of the reward signal, enabling the training of effective reward models and DPO models without requiring the full response.

\subsection{Training Reward Models and DPO Models with Truncated Preference Data}

In this work, we investigate the effects of truncating the responses in preference datasets at various positions. Let \( r_{\text{cho}}(i)^{\text{trunc}} \) and \( r_{\text{rej}}(i)^{\text{trunc}} \) denote the truncated chosen and rejected responses, respectively, where truncation is applied to retain only the first \( t_k \) tokens of each response:
$$
\begin{aligned}
r_{\text{cho}}(i)^{\text{trunc}} &= [y_1, y_2, \dots, y_{t_k}], 
\\ r_{\text{rej}}(i)^{\text{trunc}} &= [z_1, z_2, \dots, z_{t_k}]
\end{aligned}
$$

We train reward models on these truncated preference datasets. The reward model aims to predict the relative quality of responses given the truncated input. Specifically, we model the reward using the following formula:
\[
P(y \succ y' \mid x) = \frac{\exp(r(y; x))}{\exp(r(y; x)) + \exp(r(y'; x))},
\]
where \( r(y; x) \) represents the reward function for response \( y \) given the context \( x \), and \( P(y \succ y' \mid x) \) is the probability that response \( y \) is preferred over \( y' \). Although the reward model is trained on truncated responses, it is still able to assess the quality of full responses effectively by leveraging the reward function learned from the truncated portions.

Similarly, for Direct Preference Optimization (DPO), we fine-tune a base model on the truncated preference datas. The DPO objective seeks to maximize the likelihood of the chosen response over the rejected response by minimizing the following loss:
\[
\ell_{\text{DPO}}(x, y_w^{\text{trunc}}, y_l^{\text{trunc}}; \theta; \pi_{\text{ref}}) :=- \log \sigma\left( \beta \left[ \log \frac{\pi_\theta(y_w^{\text{trunc}} \mid x)}{\pi_{\text{ref}}(y_w^{\text{trunc}} \mid x)} - \log \frac{\pi_\theta(y_l^{\text{trunc}} \mid x)}{\pi_{\text{ref}}(y_l^{\text{trunc}} \mid x)} \right] \right),
\]
where \( \pi_\theta \) is the probability distribution generated by the model, \( \pi_{\text{ref}} \) is the reference model's distribution, \( y_w^{\text{trunc}} \) and \( y_l^{\text{trunc}} \) represent the truncated winning and losing responses, and \( \sigma \) is the sigmoid function. In our approach, we train the DPO model on truncated responses, but it is still capable of generating full responses and performing in regular dialogues. The truncation helps to focus on the most relevant tokens early in the response, reducing noise from irrelevant parts of the response.

\subsection{Mixing Strategy and Decoding Policies}

To further optimize the alignment between the model's output and human preferences, we propose a mixing strategy and two novel decoding policies. The mixing strategy combines the DPO policy with the corresponding reference model policy to enhance the reward-KL divergence tradeoff.

\subsubsection{Mixing Strategy}
\label{sec:mixing-strategy}

The mixing strategy involves combining the probability distributions from the DPO model \( \pi_{\text{DPO}} \) and the reference model \( \pi_{\text{ref}} \) in a weighted manner. Specifically, we define a mixing policy \( \pi_{\text{mix}} \) as:
$$
\begin{aligned}
\pi_{\text{mix}} = \text{softmax} \left( a \cdot \log \frac{\pi_{\text{DPO}}}{\pi_{\text{ref}}} + \log \pi_{\text{ref}} \right)
\end{aligned}
$$
where \( a \) is a mixing coefficient controlling the tradeoff between the DPO and reference model. This strategy allows for fine-tuning the balance between the reward signal captured by the DPO policy and the stability provided by the reference model.

\subsubsection{Decoding Strategies}
\label{sec:decoding-strategies}

We explore two decoding strategies that prioritize the early part of the response or manage the KL divergence between the DPO and reference models.

\textbf{Length Control Decoding:} In this strategy, the first \( t \) tokens are generated by sampling from the DPO policy, while the remaining tokens are generated by sampling from the reference model. The goal is to focus on the part of the response where the reward signal is concentrated. The strategy is parameterized by the truncation length \( t \), which controls the point at which the decoding switches between the two models.
$$
\begin{aligned}
y_k &= \begin{cases}
\text{sample from } \pi_{\text{DPO}} & \text{if } k \leq t \\
\text{sample from } \pi_{\text{ref}} & \text{if } k > t
\end{cases}
\end{aligned}
$$
\textbf{KL Threshold Control Decoding:} In this strategy, we compute the KL divergence between the DPO model and the reference model at each token generation step. If the KL divergence exceeds a predefined threshold \( b \), we sample from the DPO policy; otherwise, we sample from the reference model. This dynamic approach allows the model to maintain flexibility in adjusting to the relative importance of reward signal versus stability during the response generation process.
$$
\begin{aligned}
y_t &= \begin{cases}
\text{sample from } \pi_{\text{DPO}} & \text{if } \text{KL}(\pi_{\text{DPO}} \parallel \pi_{\text{ref}}) > b \\
\text{sample from } \pi_{\text{ref}} & \text{if } \text{KL}(\pi_{\text{DPO}} \parallel \pi_{\text{ref}}) \leq b
\end{cases}
\end{aligned}
$$
where \( y_t^{(i)} \) denotes the \(i\)-th sampled token from the DPO model at the \( t \)-th position.

The KL divergence \( \text{KL}(\pi_{\text{DPO}} \parallel \pi_{\text{ref}}) \) is computed at each token position as:
\[
\text{KL}(\pi_{\text{DPO}} \parallel \pi_{\text{ref}}) = \mathbb{E}_{y_t \sim \pi_{\text{DPO}}} \left[ \log \frac{\pi_{\text{DPO}}(y_t | x, y_{<t})}{\pi_{\text{ref}}(y_t | x, y_{<t})} \right]
\]

This expectation is estimated using Monte Carlo sampling. Specifically, we sample \( K = 1,000 \) tokens from the DPO model at each token position, and the KL divergence is computed as:
\[
\hat{\text{KL}}(\pi_{\text{DPO}} \parallel \pi_{\text{ref}}) = \frac{1}{K} \sum_{i=1}^{K} \log \frac{\pi_{\text{DPO}}(y_t^{(i)} | x, y_{<t})}{\pi_{\text{ref}}(y_t^{(i)} | x, y_{<t})}
\]

Both of these strategies aim to improve the reward alignment while maintaining a favorable KL divergence, leading to better model outputs.

\section{Experiment: Truncation Effects on Reward Models and DPO}
\label{sec: truncation_reward_dpo}

\subsection{Experiment Setting}

In this experiment, we investigate the effect of truncating response sequences at different positions within preference datasets Skywork-Reward-Preference-80K-v0.2 \cite{DBLP:journals/corr/abs-2410-18451}, ultrafeedback-binarized \cite{DBLP:journals/corr/abs-2310-01377}, and RLHFlow-pair-data-v2-80K-wsafety\footnote{\url{https://huggingface.co/datasets/RLHFlow/pair_data_v2_80K_wsafety}}, which are commonly used in the context of large language models. Specifically, we apply truncation to the response sections (including both chosen and rejected responses) at varying positions. The truncation process retains only the initial portion of the response tokens, while the remaining tokens are discarded, resulting in the creation of multiple truncated datasets. We then train reward models and use Direct Preference Optimization (DPO) to fine-tune models on these truncated datasets and compare their performance with models trained on the original, untruncated datasets. We also investigate the use of DPO implicit reward~\cite{DBLP:conf/nips/RafailovSMMEF23} to assess the quality of two responses on datasets with different truncation ratios, and compare the accuracy of this evaluation with the actual quality judgments.

We utilize Google's gemma-2b-it\footnote{\url{https://huggingface.co/google/gemma-2b-it}} model as the base for training the reward model, following the methodology outlined in RLHFlow~\cite{dong2024rlhf} to train a standard Bradley-Terry reward model~\cite{Bradley1952RankAO}. For the DPO training, we use the Llama-3.1-8B-Instruct~\cite{DBLP:journals/computer/PattersonGHLLMR22} as the base model, following the DPO methodology outlined in OpenRLHF~\cite{hu2024openrlhf} to fine-tune the model. In the experiment using DPO implicit reward to assess accuracy, we use the LLaMA3-iterative-DPO-final model~\cite{xiong2024iterative, dong2024rlhf} as the DPO policy model and its supervised fine-tuning (SFT) checkpoint, LLaMA3-SFT, trained from Llama-3-8B, as the reference policy model. All experiments are performed using 80GB A100 or H100 GPUs

\subsubsection{Metrics}

The performance of the models is evaluated using two metrics:

\noindent \textbf{Test Accuracy}. This metric measures the proportion of instances where the reward model assigns a higher score to the chosen response compared to the rejected response.

\noindent \textbf{GPT4o Win Rate}. This metric is computed using the AlpacaEval 2.0~\cite{alpaca_eval} standard test set and the default baseline model with GPT4o acting as the judge.

\subsection{Results}

\subsubsection{Evaluation of Reward Models on RewardBench}
\label{subsec:rewardbench} 

We evaluate the performance of the trained reward models on the core RewardBench evaluation set. For each dataset, we train the reward models on the training set using truncated versions of the responses with truncation ratios of 50\%, 40\%, 33\% and 25\%. The results are presented in \autoref{tab:rewardbench_results}. 

\begin{table*}[h]
\centering
\begin{tabular}{ccccccc}
\hline
Dataset & Dimension & Original Dataset & 50\% & 40\% & 33\% & 25\% \\
\cline{1-7}
\multirow{5}{*}{\textbf{Skywork-Preference}}   
 & Chat & \cellcolor{bestblue}0.8073 & \cellcolor{secondblue}0.7318 & 0.7039 & 0.5866 & 0.5978 \\
 & Chat-Hard & \cellcolor{secondblue}0.7039 & \cellcolor{bestblue}0.7105 & 0.6974 & 0.6776 & 0.6732 \\
 & Safety & \cellcolor{bestblue}0.8216 & 0.8068 & 0.7946 & \cellcolor{secondblue}0.8162 & 0.8030 \\
 & Reasoning & 0.7043 & \cellcolor{secondblue}0.7769 & \cellcolor{bestblue}0.8101 & 0.7064 & 0.7450 \\
 & Total & 0.7585 & \cellcolor{secondblue}0.7588 & \cellcolor{bestblue}0.7635 & 0.7000 & 0.6992 \\
\hline
\multirow{5}{*}{\textbf{UltraFeedback}}   
 & Chat & 0.7946 & \cellcolor{bestblue}0.8098 & \cellcolor{secondblue}0.8073 & 0.7844 & 0.7644 \\
 & Chat-Hard & 0.6029 & \cellcolor{bestblue}0.6425 & \cellcolor{secondblue}0.6342 & 0.5983 & 0.5946 \\
 & Safety & 0.7416 & \cellcolor{secondblue}0.7632 & \cellcolor{bestblue}0.7848 & 0.7384 & 0.6756 \\
 & Reasoning & \cellcolor{bestblue}0.7056 & \cellcolor{secondblue}0.6904 & 0.6682 & 0.6886 & 0.5646 \\
 & Total & \cellcolor{bestblue}0.7391 & \cellcolor{secondblue}0.7327 & 0.7194 & 0.7018 & 0.6355 \\
\hline
\multirow{5}{*}{\textbf{RLHFlow-Preference}}   
 & Chat & \cellcolor{bestblue}0.9553 & \cellcolor{secondblue}0.9302 & 0.9287 & 0.8574 & 0.8291 \\
 & Chat-Hard & \cellcolor{secondblue}0.4517 & \cellcolor{bestblue}0.4561 & 0.4506 & 0.4323 & 0.4127 \\
 & Safety & \cellcolor{bestblue}0.6730 & \cellcolor{secondblue}0.6621 & 0.6438 & 0.5985 & 0.6081 \\
 & Reasoning & 0.5984 & \cellcolor{bestblue}0.8374 & \cellcolor{secondblue}0.7894 & 0.6247 & 0.5723 \\
 & Total & 0.6596 & \cellcolor{bestblue}0.7216 & \cellcolor{secondblue}0.6971 & 0.6244 & 0.5562 \\
\hline

\end{tabular}
\caption{Performance of reward models trained on different truncation ratios for various datasets. The table presents the evaluation scores across multiple dimensions from the RewardBench core set: \textbf{Chat}, \textbf{Chat-Hard}, \textbf{Safety} and \textbf{Reasoning}. \textbf{Total} is the final score on the RewardBench core set.
\textbf{Skywork-Preference} refers to Skywork-Reward-Preference-80K-v0.2 dataset, \textbf{UltraFeedback} refers to ultrafeedback-binarized dataset, \textbf{RLHFlow-Preference} refers to RLHFlow-pair-data-v2-80K-wsafety dataset.
\textbf{Original Dataset} refers to the model trained on the full dataset without truncation;
\textbf{50\%}, \textbf{40\%}, \textbf{33\%}, and \textbf{25\%} refer to truncated datasets with corresponding ratios. The highest score in each row is highlighted with \colorbox{bestblue}{\textcolor{black}{darker blue}}, and the second-highest score with \colorbox{secondblue}{\textcolor{black}{lighter blue}}.
}
\label{tab:rewardbench_results}
\end{table*}

Truncating the response in the preference data to 50\% or 40\% of tokens had minimal impact on the performance of the trained reward model across all three datasets. In fact, for certain metrics and datasets, models trained on truncated data outperformed those trained on full responses. However, truncating the response to 33\% or 25\% of its original length leads to a slight reduction in performance. Despite this, the performance drop remains small, and the models continue to exhibit the majority of the performance seen with the original, untruncated datasets.

\subsubsection{Evaluation of Reward Models on Each Task of UltraFeedback}

We train reward models on the ultrafeedback-binarized dataset, separately for each task: Helpfulness, Honesty, Instruction Following, and Truthfulness. For each task, we train the reward models on the training set using truncated versions of the responses with truncation ratios of 50\%, 40\%, 30\%, 20\% and 10\%. Results are shown in \autoref{tab:ultrafeedback_accuracy}. 

The results show that truncating the responses to 50\% or 40\% of their original length had a negligible effect on test accuracy for each task. In some tasks, models trained on truncated data even perform better than those trained on full responses. However, when the responses are truncated to shorter lengths (e.g., 30\%, 20\%, or 10\%), a slight decrease in test accuracy is observed. Nonetheless, the models retain a substantial portion of their original performance, indicating that truncation did not result in a significant loss of accuracy.

\begin{table*}[ht]
\centering
\begin{tabular}{ccccccc}
\hline
{Task} & {Original Dataset} & {50\%} &{40\% } &{30\% } &{20\% } & {10\%} \\
\hline
Helpfulness & 0.89 & \cellcolor{bestblue}0.90 & \cellcolor{bestblue}0.90 & 0.87 & 0.82 & 0.73 \\
Honesty & \cellcolor{secondblue}0.87 & \cellcolor{bestblue}0.88 & \cellcolor{secondblue}0.87 & 0.84 & 0.79 & 0.76 \\
Instruction Following & \cellcolor{bestblue}0.91 & \cellcolor{bestblue}0.91 & 0.86 & 0.87 & 0.74 & 0.69 \\
Truthfulness & \cellcolor{bestblue}0.85 & \cellcolor{secondblue}0.84 & \cellcolor{secondblue}0.84 & 0.83 & 0.81 & 0.64 \\
Average & \cellcolor{secondblue}0.88 & \cellcolor{bestblue}0.8825 & 0.87 & 0.855 & 0.795 & 0.705 \\
\hline

\end{tabular}
\caption{UltraFeedback test accuracy across different tasks with various truncation ratios. The table presents the test accuracy for each task in the UltraFeedback dataset, with different truncation ratios: \textbf{Original Dataset} refers to the model evaluated on the full, unmodified UltraFeedback dataset; 
\textbf{50\%}, \textbf{40\%}, \textbf{30\%}, \textbf{20\%}, and \textbf{10\%} refer to models evaluated using truncated versions of the dataset. The tasks listed include: \textbf{Helpfulness}, \textbf{Honesty}, \textbf{Instruction Following}, and \textbf{Truthfulness}. \textbf{Average} represents the mean accuracy across all tasks.
The highest score in each row is highlighted with \colorbox{bestblue}{\textcolor{black}{darker blue}}, and the second-highest score with \colorbox{secondblue}{\textcolor{black}{lighter blue}}.
}
\label{tab:ultrafeedback_accuracy}
\end{table*}

\subsubsection{Evaluation of DPO-trained Models on AlpacaEval 2.0}

In addition to training reward models, we investigate the effect of response truncation in the preference dataset by Direct Preference Optimization (DPO). For this experiment, we use the Skywork-Reward-Preference-80K-v0.2 dataset~\cite{DBLP:journals/corr/abs-2410-18451}. The dataset responses are truncated at various ratios of 50\%, 40\%, 33\% and 25\%. Results are shown in \autoref{tab:alpaca_dpo}.

The results indicate that truncating the responses in the preference data had a minimal effect on the performance of models trained with DPO. While the impact increased with the truncation ratio, truncating the response to 50\% or 40\% of its original length does not significantly degrade the performance of the DPO-trained models. This suggests that, in the context of DPO training, the majority of the signals used to evaluate response quality are concentrated in the earlier segments of the response.

\begin{table*}[ht]
\centering
\begin{tabular}{ccccccc}
\hline
{Metric} & {Llama3.1 8B} & {Original Dataset} & {50\%} & {40\%} & {33\%} & {25\%} \\
\hline
LCWR &  21.45 & \cellcolor{secondblue}24.90 & \cellcolor{bestblue}25.19 & 24.85 & 23.51 & 21.13 \\
WR   &  22.37 & \cellcolor{secondblue}23.92 & \cellcolor{bestblue}24.15 & 23.57 & 23.43 & 20.96 \\
\hline

\end{tabular}
\caption{Performance of DPO models with different truncation ratios. The table presents the evaluation metrics for both the original model and the DPO models trained on truncated datasets:
\textbf{Llama3.1 8B} refers to the original Llama-3.1-8B-Instruct model;
\textbf{Original Dataset} refers to the Llama-3.1-8B-Instruct model fine-tuned using the full Skywork-Reward-Preference-80K-v0.2 dataset with the DPO algorithm;
\textbf{50\%}, \textbf{40\%}, \textbf{33\%}, and \textbf{25\%} refer to models fine-tuned using truncated versions of the dataset.
\textbf{LCWR} refers to Length-controlled Win Rate and
\textbf{WR} refers to Win Rate. The highest score in each row is highlighted with \colorbox{bestblue}{\textcolor{black}{darker blue}}, and the second-highest score with \colorbox{secondblue}{\textcolor{black}{lighter blue}}.}
\label{tab:alpaca_dpo}
\end{table*}

\subsubsection{Implicit Reward Accuracy on Truncated Responses}
\label{subsec:implicit-reward-accuracy}

\begin{figure}[htbp]
    \centering
    \subfloat[Predict Accuracy vs Truncated Ratio]{
        \includegraphics[width=0.45\linewidth]{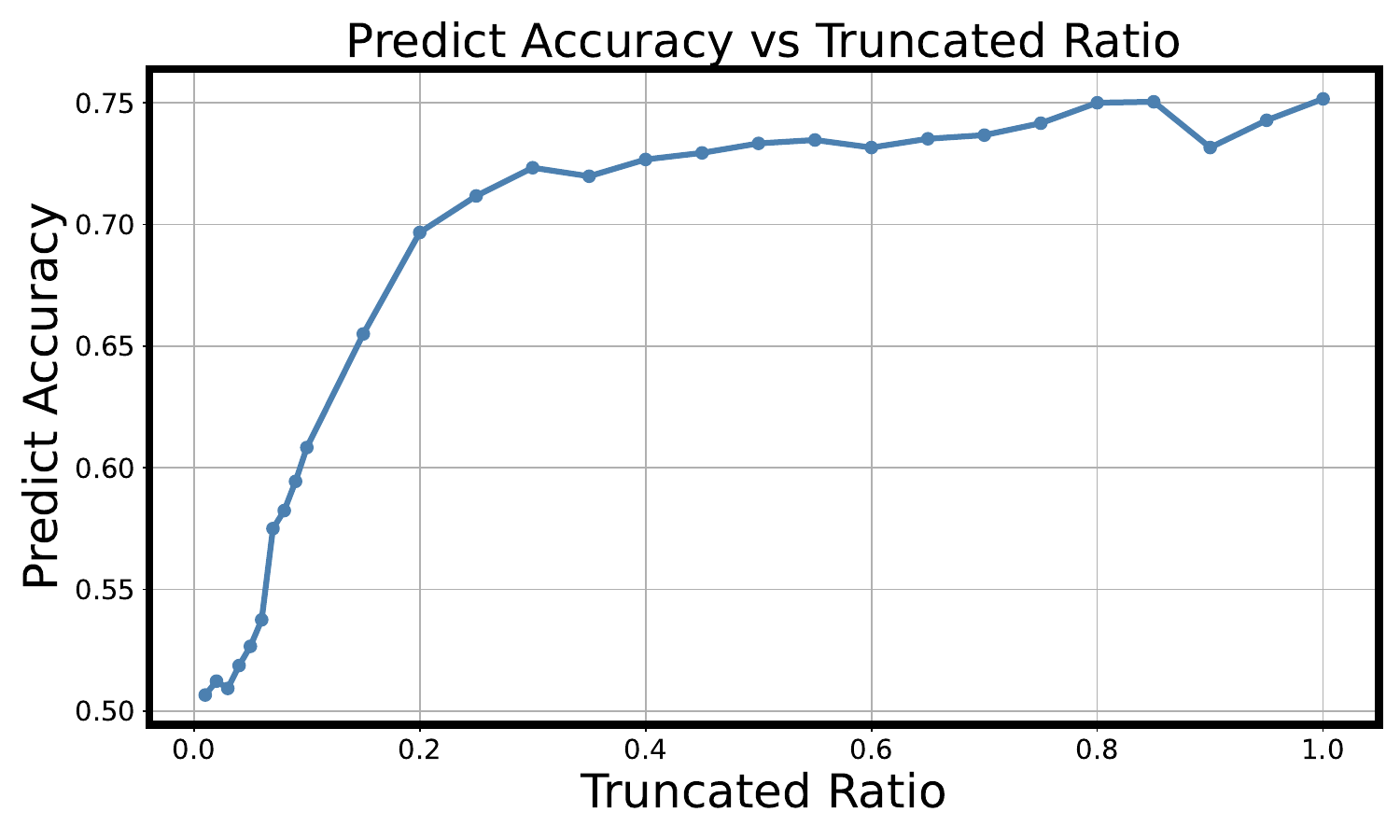}
        \label{fig:dpo_implicit_reward_accuracy_vs_ratio}
    }
    \hspace{0mm}
    \subfloat[Predict Accuracy vs Truncated Length]{
        \includegraphics[width=0.45\linewidth]{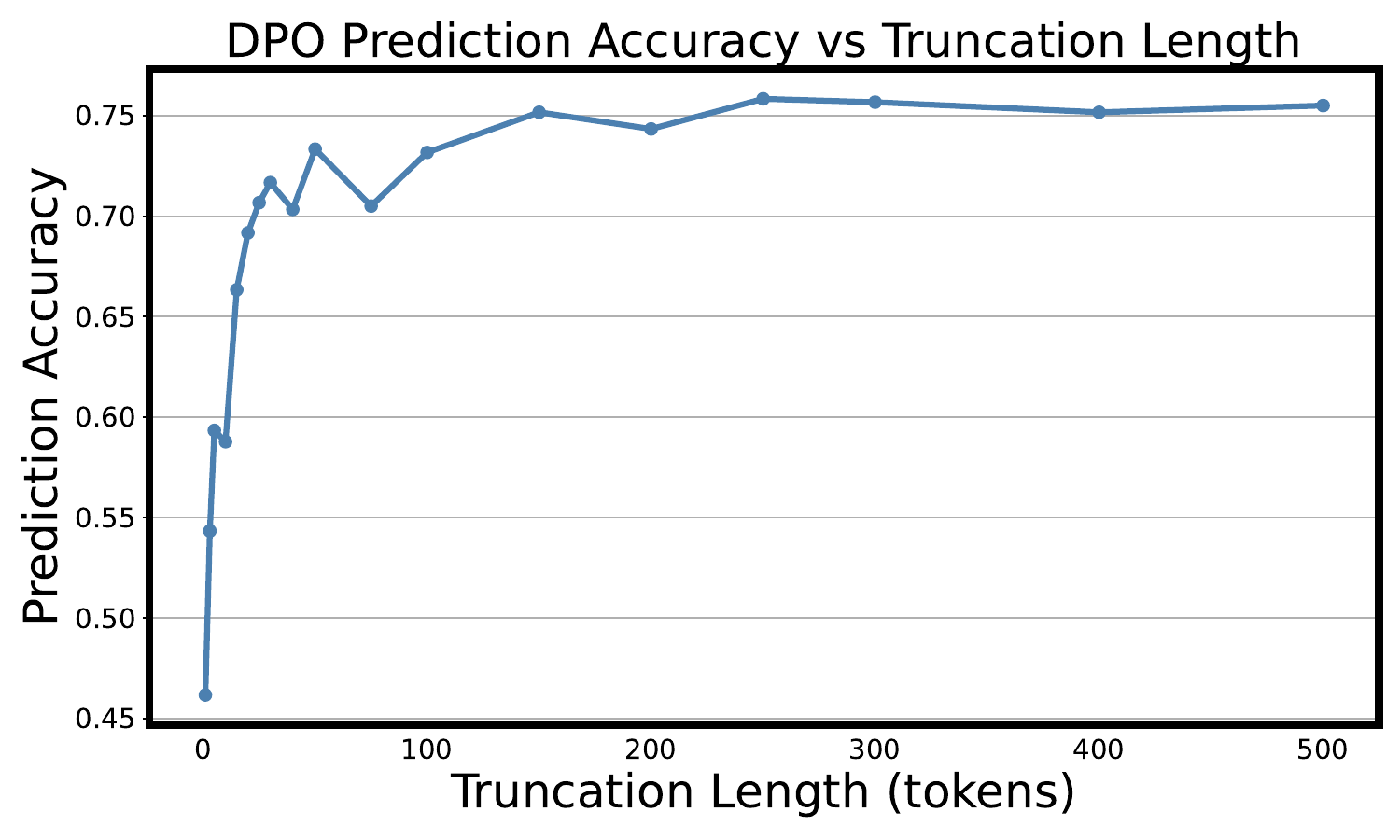}
        \label{fig:dpo_implicit_reward_accuracy_vs_length}
    }
    \caption{The x-axis represents the response truncation length and ratio, while the y-axis shows the accuracy of DPO implicit reward in predicting the relative quality of responses based on truncated datasets.}
    \label{fig:dpo_implicit_reward_accuracy}
\end{figure}

In this experiment, we truncate the responses in the Skywork-Reward-Preference-80K-v0.2~\cite{DBLP:journals/corr/abs-2410-18451} dataset at various proportions and compute the DPO implicit reward for each response pair. We then compare the preferences derived from the implicit rewards with the actual human-annotated preferences to assess the consistency. The results are presented in \autoref{fig:dpo_implicit_reward_accuracy}.

The results indicate that as the length of the response considered increases, the preferences derived from the DPO implicit reward align more closely with human-annotated preferences. Interestingly, even when only the initial portion of the response is considered, the preferences derived from the DPO implicit reward show a high degree of consistency with human preferences. This suggests that, in preference datasets, evaluating only the early tokens of a response is sufficient to accurately assess the relative quality of two responses, without the need to examine the entire response.

\section{Experiment: KL Divergence and Reward-KL Tradeoff for Evaluating Response Quality}
\label{sec:kl-divergence-experiment}

This section presents a set of experiments that examine the relationship between the Kullback-Leibler (KL) divergence between the DPO model and the reference model, and the reward-KL tradeoff during response generation. These experiments aim to validate the hypothesis that the reward signal in preference datasets is primarily concentrated in the early part of the response.

\subsection{Experiment Setup}
To investigate this hypothesis, we perform two key experiments. In the first experiment, we compute the KL divergence between the DPO model and the reference model at each token generation step. This experiment allows us to observe how the KL divergence evolves as the response is generated and whether the early tokens exhibit a higher divergence compared to later ones. In the second experiment, we explore the reward-KL tradeoff during generation by adjusting the sampling strategy based on the DPO model and reference model, to further confirm the concentration of reward signal in the early part of the response. We use the mixing decoding strategy described in ~\autoref{sec:mixing-strategy} under different $a$ as a baseline, and test different decoding strategies. 

For both experiments, we use the LLaMA3-iterative-DPO-final model~\cite{xiong2024iterative, dong2024rlhf} as the DPO policy model and its supervised fine-tuning (SFT) checkpoint, LLaMA3-SFT, trained from Llama-3-8B, as the reference policy model. And we measure the corresponding reward using the reward model FsfairX-LLaMA3-RM-v0.1~\cite{dong2024rlhf}. We randomly selected 1000 instructions from the training sets of Alpaca~\cite{alpaca} and UltraFeedback~\cite{DBLP:journals/corr/abs-2310-01377} to form the instruction sets for these two experiments. The KL divergence between two policy at a token is computed as described in \autoref{sec:decoding-strategies} and the KL divergence between two policy for the whole response generation is accumulated across all token generation steps, yielding the final KL divergence as follows:
\[
\hat{\text{KL}}(\pi_{\text{mix}} \parallel \pi_{\text{ref}}) = \frac{1}{N} \sum_{i=1}^{N} \sum_{t=1}^{T} \log \frac{\pi_{\text{mix}}(y_t^{(i)} | x_i, y_{<t})}{\pi_{\text{ref}}(y_t^{(i)} | x_i, y_{<t})}
\]
where \( N \) represents the size of the instruction set, \( T \) denotes the total number of tokens in the response, \( x_i \) is the instruction, \( y_t^{(i)} \) refers to the generated token at position \( t \), and \( y_{<t} \) refers to the tokens generated prior to token \( t \).

\subsection{Results}
\subsubsection{KL Divergence Analysis Across Token Positions}

\begin{figure}[ht]
\centering
\includegraphics[width=0.5\textwidth]{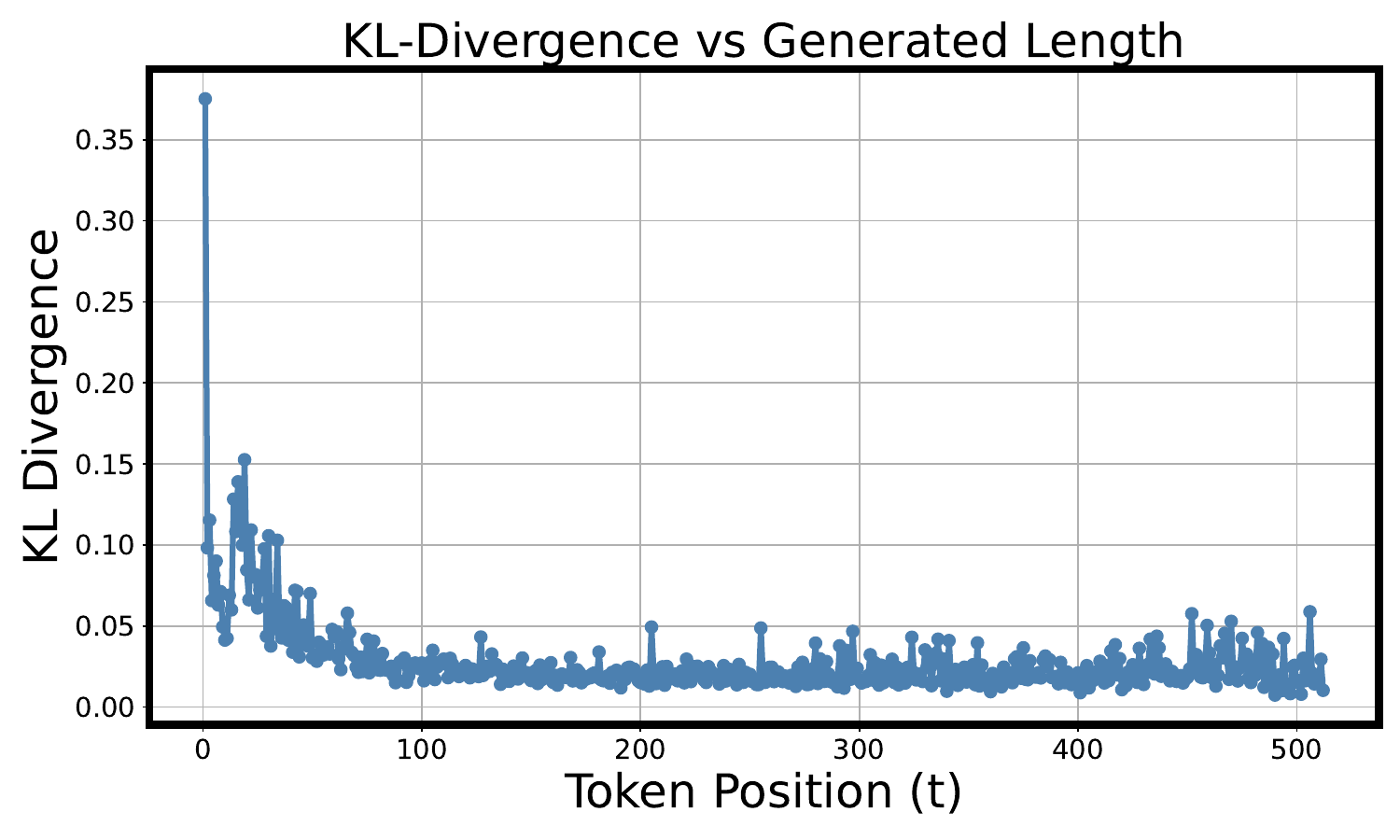}
\caption{KL Divergence between the DPO model and the reference model at each token position. The plot shows that the divergence is higher for early tokens and decreases as generation progresses.}
\label{fig:kl-divergence}
\end{figure}

In the first experiment, we analyze the KL divergence between the DPO model and the reference model at each token generation step. The KL divergence is computed for each token \( y_t \) by comparing the conditional probability distributions of the DPO model \( \pi_{\text{DPO}}(y_t | x, y_{<t}) \) and the reference model \( \pi_{\text{ref}}(y_t | x, y_{<t}) \), where \( x \) is the instruction, and \( y_{<t} \) represents previously generated tokens. As shown in Figure~\ref{fig:kl-divergence}, the KL divergence is high in the early tokens, indicating significant differences between the DPO and reference models. However, the divergence diminishes significantly as token generation progresses, suggesting that the primary divergence occurs in the initial phase of response generation.

This observation supports the hypothesis that the reward signal in preference datasets is mostly concentrated in the first part of the response, with minimal divergence in the later tokens, where the DPO model relies on the tokens generated earlier.

\subsubsection{Reward-KL Tradeoff for Length Control and KL Threshold Control Decoding}

The second experiment explores the reward-KL tradeoff during response generation with different decoding strategies. We focus on two strategies: Length Control Decoding and KL Threshold Control Decoding as described in~\autoref{sec:decoding-strategies} and evaluate them under different parameters. 

\paragraph{Length Control Decoding}
In Length Control Decoding, we sample from the DPO policy for the first \( t \) tokens and from the reference policy for the remaining tokens. We evaluate this strategy for various values of \( t \) and compute the average reward and KL divergence for each configuration.

\paragraph{KL Threshold Control Decoding}
In KL Threshold Control Decoding, we compute the KL divergence \( \text{KL}(\pi_{\text{DPO}} \parallel \pi_{\text{ref}}) \) at each token position. If the divergence exceeds a threshold \( b \), we sample from the DPO policy; otherwise, we sample from the reference policy. We test several values of \( b \) and record the average reward and KL divergence.

\begin{figure}[ht]
\centering
\includegraphics[width=0.5\textwidth]{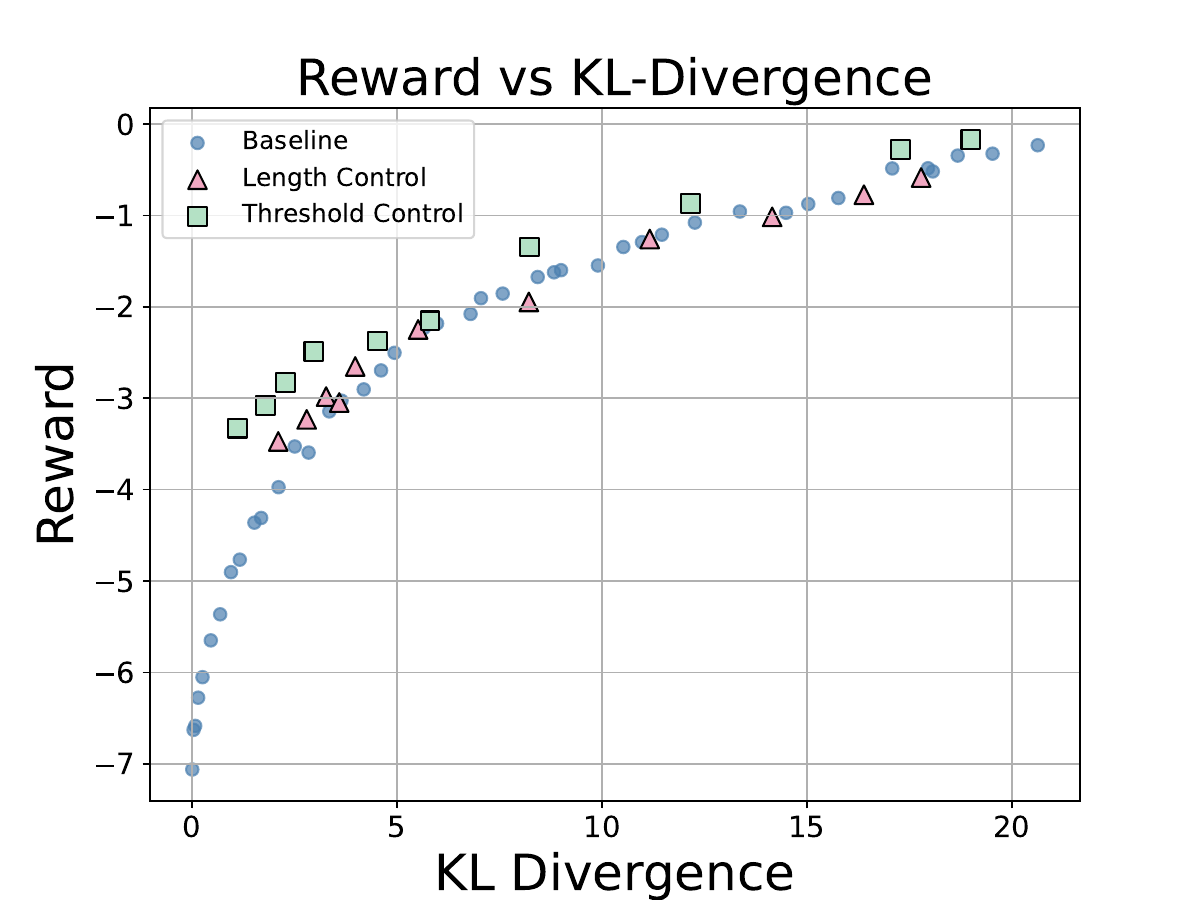}
\caption{Reward and corresponding KL Divergence for the baseline and two different control strategies. The blue dots represent data from the baseline, while the red triangles and green squares represent the Length Control and KL Threshold Control strategies, respectively.}
\label{fig:reward_vs_kl_div}
\end{figure}

The results of both strategies, shown in Figure~\ref{fig:reward_vs_kl_div}, demonstrate that both decoding strategies improve the reward-KL tradeoff compared to the baseline. These findings confirm that adjusting the decoding strategy based on the KL divergence between the DPO and reference models leads to a better alignment between reward and KL divergence, further supporting the idea that the reward signal is concentrated in the early tokens.

\subsection{Discussion}
These experiments validate the hypothesis that the reward signal used in the DPO model is concentrated in the early part of the response. The analysis of KL divergence reveals that the primary differences between the DPO and reference models occur in the initial token generation, while the reward-KL tradeoff experiments demonstrate how adjusting the sampling strategy can improve the alignment between reward and KL divergence. These findings highlight the importance of the early response tokens in shaping the overall quality of generated responses.

\section{Investigating the Autoregressive Influence on Preference Signals}
\label{sec:autoregressive_influence}

In previous experiments, we observed that the preference signal appears to be concentrated in the initial portion of the response sequence. This could potentially be an artifact of the autoregressive nature of the data generation process. Given that the datasets used in earlier experiments were synthesized using autoregressive language models, we hypothesize that this phenomenon might be influenced by the autoregressive paradigm itself. 

To validate this hypothesis, we conducted a series of experiments using human-generated responses and preference labels. Specifically, we employed the SHP dataset~\cite{pmlr-v162-ethayarajh22a}, which consists of responses and preference annotations generated by humans, to repeat the experiments outlined in~\autoref{subsec:rewardbench} and~\autoref{subsec:implicit-reward-accuracy}. 

\subsection{Results}

\subsubsection{Performance on RewardBench}

We trained reward models on the human-generated SHP dataset using both original and truncated versions of the responses. The evaluation was conducted on the RewardBench core set. The results, shown in~\autoref{tab:shp_rewardbench_results}, demonstrate that the shallow preference signal phenomenon persists even when using human-generated data. 

\begin{table*}[h]
\centering
\begin{tabular}{ccccccc}
\hline
Dataset & Dimension & Original Dataset & 50\% & 40\% & 33\% & 25\% \\
\cline{1-7}
\multirow{5}{*}{\textbf{SHP-Preference}}   
 & Chat & \cellcolor{bestblue}{0.8198} & 0.8071 & \cellcolor{secondblue}{0.8139} & 0.7874 & 0.7709 \\
 & Chat-Hard & \cellcolor{secondblue}{0.6039} & \cellcolor{bestblue}{0.6352} & 0.5759 & 0.5155 & 0.5274 \\
 & Safety & \cellcolor{secondblue}{0.7906} & \cellcolor{bestblue}{0.8049} & 0.7825 & 0.7698 & 0.7589 \\
 & Reasoning & \cellcolor{bestblue}{0.5624} & 0.5532 & 0.5439 & \cellcolor{secondblue}{0.5592} & 0.5451 \\
 & Total & \cellcolor{secondblue}{0.7008} & \cellcolor{bestblue}{0.7056} & 0.6989 & 0.6882 & 0.6712 \\
\hline
\end{tabular}
\caption{Performance of reward models trained on the human-generated SHP dataset with different truncation ratios. The results show the evaluation scores across multiple dimensions: \textbf{Chat}, \textbf{Chat-Hard}, \textbf{Safety}, \textbf{Reasoning}, and \textbf{Total}. \textbf{Original Dataset} refers to the model trained on the full dataset without truncation; \textbf{50\%}, \textbf{40\%}, \textbf{33\%}, and \textbf{25\%} refer to datasets where the responses are truncated to retain 50\%, 40\%, 33\%, and 25\% of the original token length, respectively. The highest score in each row is highlighted with \colorbox{bestblue}{\textcolor{black}{darker blue}}, and the second-highest score with \colorbox{secondblue}{\textcolor{black}{lighter blue}}.}
\label{tab:shp_rewardbench_results}
\end{table*}

\subsubsection{DPO Implicit Reward Accuracy on Human-Generated Data}

\begin{figure}[htbp]
    \centering
    \subfloat[Predict Accuracy vs Truncated Ratio]{
        \includegraphics[width=0.45\linewidth]{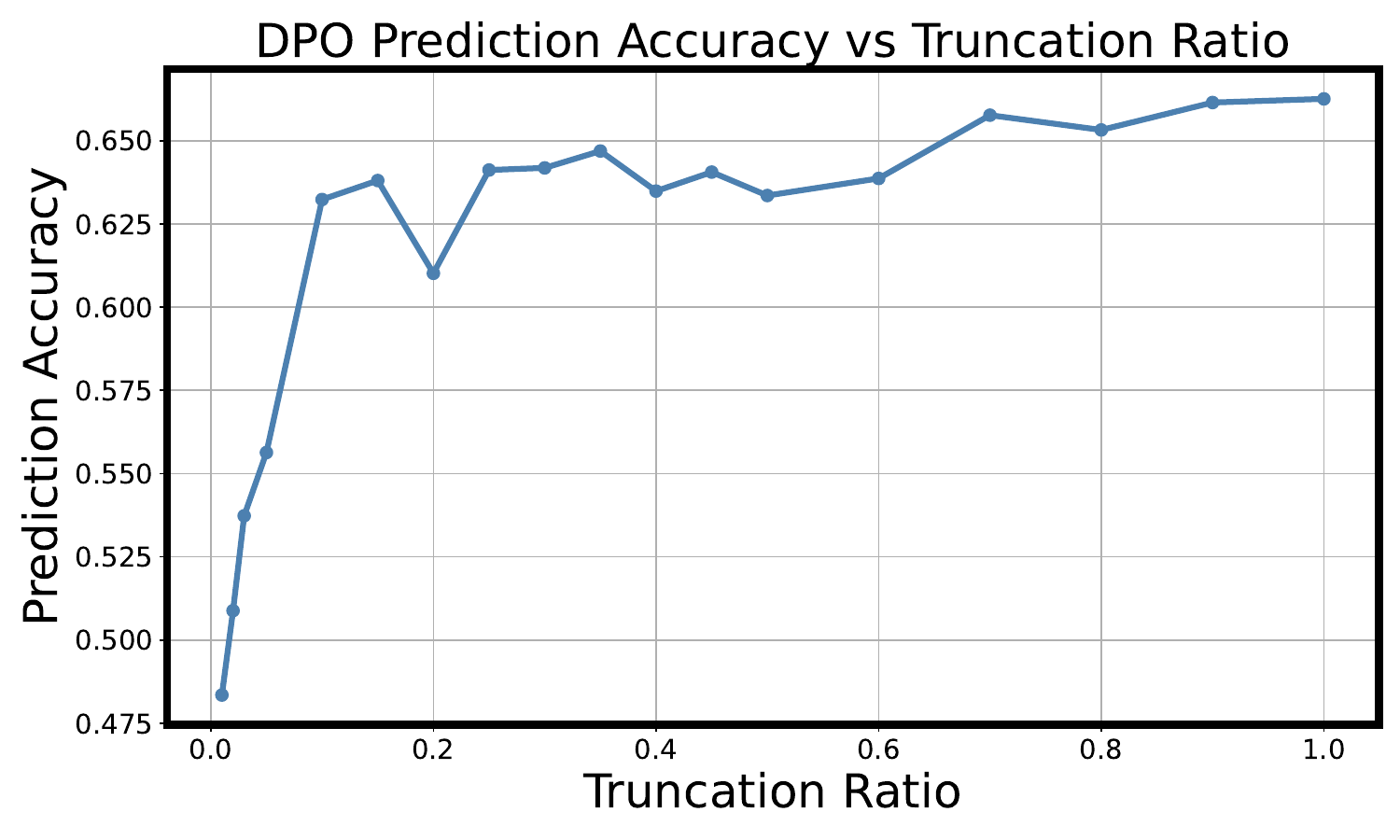}
        \label{fig:shp_generated_accuracy_vs_ratio}
    }
    \hspace{0mm}
    \subfloat[Predict Accuracy vs Truncated Length]{
        \includegraphics[width=0.45\linewidth]{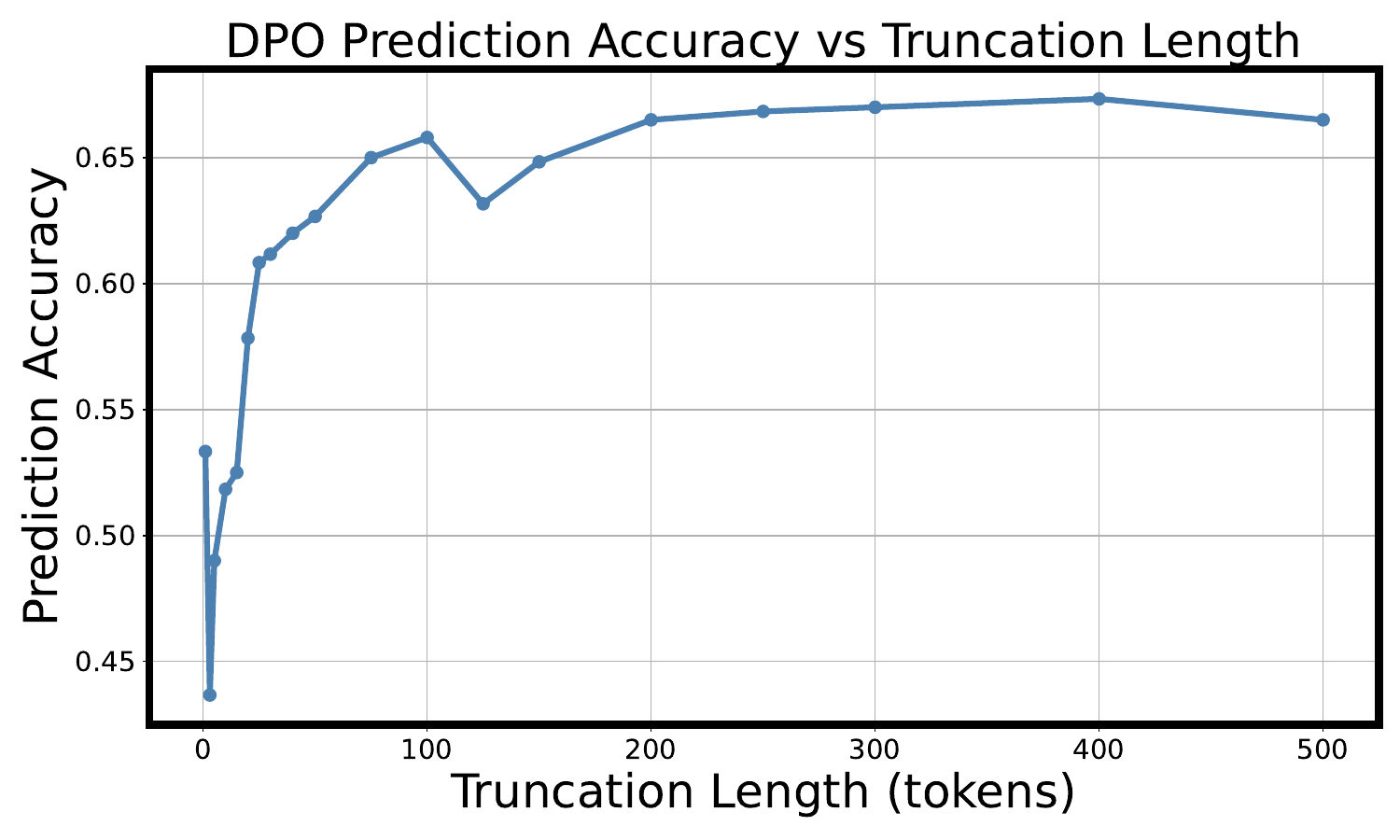}
        \label{fig:shp_generated_accuracy_vs_length}
    }
    \caption{Accuracy of DPO implicit reward in predicting the relative quality of responses on the human-generated SHP dataset with truncated responses. The x-axis represents the truncation ratio and length, and the y-axis shows the accuracy of DPO implicit reward predictions compared to human annotations.}
    \label{fig:shp_generated_accuracy}
\end{figure}

We also applied the DPO implicit reward approach to the truncated human-generated responses, as described in~\autoref{subsec:implicit-reward-accuracy}, to predict the relative quality of response pairs. The accuracy of these predictions was then compared to human-annotated preferences. The results, shown in~\autoref{fig:shp_generated_accuracy}, confirm that the shallow preference signal phenomenon persists even with human-generated data. As the truncation ratio decreases, the alignment between DPO implicit reward predictions and human-annotated preferences remains high, demonstrating that even truncated responses are sufficient for accurately predicting relative quality.

\subsection{Conclusion}

The results from the human-generated data experiments provide strong evidence that the observed shallow preference signal is not solely a byproduct of autoregressive data generation. Even when the data is generated by humans, the preference signal remains concentrated in the early portions of the response. This indicates that the phenomenon is likely inherent in the structure of the response itself, rather than an artifact of the autoregressive generation process.

\section{Limitations}
\label{sec:limitations}



One limitation of this work is the absence of a strong theoretical foundation for the proposed phenomenon. Although our empirical results are compelling, a comprehensive theoretical explanation of the specific parts of a response that contribute to human preferences remains elusive. Future research could explore this aspect in more depth to establish a more robust theoretical framework.

\section{Conclusion}
\label{sec:conclusion}

We introduce shallow preference signals, where key distinguishing features between preferred and non-preferred responses are concentrated in early response tokens. Our experiments show that models trained on truncated data—retaining 40\% to 50\% of tokens—perform similarly or better in reward modeling and Direct Preference Optimization than those trained on full-length data. Additionally, we highlight the limitation of current methods that focus mainly on initial tokens, suggesting the need for strategies that consider entire responses for more accurate alignment with human preferences.

\bibliographystyle{unsrt}  
\bibliography{templateArxiv}  

\appendix

\section{Preliminaries}
\label{sec:preliminaries}


\subsection{Autoregressive Language Model And Token-Level Markov Decision Process}

Autoregressive language models (ARLMs) are designed to generate token sequences \( y_1, y_2, \dots, y_T \) conditioned on the preceding tokens in a given context. Formally, for a provided input prompt \( x \), the model generates the token sequence \( y = (y_1, y_2, \dots, y_T) \) by factorizing the joint distribution of the sequence using the chain rule of probability:
\[
p(y|x) = \prod_{t=1}^{T} p(y_t | y_1, y_2, \dots, y_{t-1}, x),
\]
where \( p(y_t | y_1, y_2, \dots, y_{t-1}, x) \) represents the conditional probability of generating token \( y_t \), given all previous tokens \( y_1, y_2, \dots, y_{t-1} \) and the input prompt \( x \).

This process is typically framed as a token-level Markov Decision Process (MDP), where each state at time step \( t \), denoted \( s_t \), represents the sequence of tokens generated up to that point:
\[
s_t = (x, y_1, y_2, \dots, y_{t-1}),
\]
and the action \( a_t \) corresponds to the generation of the next token \( y_t \). The transitions between states are deterministic and are given by:
\[
s_{t+1} = (x, y_1, y_2, \dots, y_t),
\]
as each subsequent state is determined solely by the previous state and the action of generating the next token.

This token-level MDP formulation is useful for various applications, such as in training RL-based models where the language model needs to learn to generate tokens that not only fit the linguistic context but also satisfy some predefined quality criteria. Moreover, recent advancements in reinforcement learning from human feedback (RLHF) have sought to fine-tune such models to align with human preferences, making this framework essential for ensuring that ARLMs produce high-quality, aligned outputs.

In the context of reinforcement learning (RL), the task is framed as a Max-Entropy RL problem, where the reward is a combination of a task-specific reward function and a regularization term. The objective is to maximize the expected sum of the rewards, along with the entropy of the policy to promote exploration:
\[
\begin{aligned}
\mathbb{E}_{x \sim X, y \sim \pi(\cdot | x)} \left[ r(y | x) + \beta \log \pi_{\text{ref}}(y | x) \right] + \beta \mathbb{E}_{x \sim X} [H(\pi(\cdot | x))] \\
\end{aligned}
\]
where \( r(y | x) \) represents the reward for generating a sequence \( y \) given the input prompt \( x \), \( \pi_{\text{ref}}(y | x) \) is a reference policy that can be used to encourage alignment with desired behaviors, and \( H(\pi(\cdot | x)) \) is the entropy of the policy at time \( t \), promoting exploration by discouraging deterministic behaviors.

At the token level, the RL objective can be rewritten as:
\[
\begin{aligned}
\mathbb{E}_{s_0 \sim X, a_t \sim \pi(\cdot | s_t)} \left[ \sum_{t=1}^{T} r'(s_t, a_t) \right] +  \beta \mathbb{E}_{s_0 \sim X} [H(\pi(\cdot | s_0))] , 
\end{aligned}
\]
where \( r'(s_t, a_t) \) is the token-level reward, defined as:
\[
r'(s_t, a_t) =
\begin{cases}
\beta \log \pi_{\text{ref}}(a_t | s_t), \text{  if } s_{t+1} \text{ is not terminal}, \\
r(y | x) + \beta \log \pi_{\text{ref}}(a_t | s_t), \text{otherwise}.
\end{cases}
\]
In this formulation, the reward function \( r(y | x) \) typically measures how well the generated sequence aligns with the desired outcome, while the entropy term \( \beta \log \pi_{\text{ref}}(a_t | s_t) \) encourages diversity in the generated tokens.

The objective in reinforcement learning is to find an optimal policy \( \pi^* \) that maximizes the expected cumulative reward. This is done by solving for the optimal Q-function \( Q^*(s_t, a_t) \), which provides the expected future reward for taking action \( a_t \) from state \( s_t \):
\[
Q^*(s_t, a_t) = r'(s_t, a_t) + V^*(s_{t+1}),
\]
where \( V^*(s_t) \) is the optimal state-value function, representing the expected reward from state \( s_t \). The optimal policy \( \pi^* \) satisfies the following equation:
\[
\beta \log \frac{\pi^*(a_t | s_t)}{\pi_{\text{ref}}(a_t | s_t)} = Q^*(s_t, a_t) - V^*(s_t).
\]
When \( t < T \), the optimal policy maximizes the difference between the state-value function of the next state and the current state, encouraging the model to generate the sequence that leads to the highest cumulative reward.

\subsection{RLHF with Reward Models}

Reinforcement learning from human feedback (RLHF) is an approach where a reward model is used to guide the training of the language model. The reward model \( r(y|x) \) evaluates the quality of a generated response \( y \) given a prompt \( x \). The goal is to maximize the expected reward by adjusting the model's parameters using a policy optimization algorithm such as Proximal Policy Optimization (PPO)

Initially, \cite{DBLP:conf/nips/ChristianoLBMLA17} proposed learning a reward model using the Bradley-Terry model to assign a score to each response. For a pair of responses \( y \) and \( y' \), the Bradley-Terry model defines the probability that \( y \) is preferred over \( y' \) as:
\[
P(y \succ y' | x) = \frac{\exp(r(y; x))}{\exp(r(y; x)) + \exp(r(y'; x))},
\]
The reward function is learned by maximizing the log-likelihood of preference predictions.

For a triplet \( (x, y_w, y_l) \), where \( y_w \) is the winner and \( y_l \) is the loser, the Direct Preference Optimization (DPO) loss is derived as follows:
\[
\ell_{\text{DPO}}(x, y_w, y_l; \theta; \pi_{\text{ref}}) := 
\]
\[
- \log \sigma\left( \beta \left[ \log \frac{\pi_\theta(y_w | x)}{\pi_{\text{ref}}(y_w | x)} - \log \frac{\pi_\theta(y_l | x)}{\pi_{\text{ref}}(y_l | x)} \right] \right),
\]
where \( \sigma(\cdot) \) is the logistic function, \( \sigma(z) = \frac{1}{1 + \exp(-z)} \) and \( \beta \) is a hyperparameter that controls the importance of the preference signal in the optimization process. This DPO method provides a more efficient and stable solution compared to traditional methods that require separate reward modeling and policy optimization.

\end{document}